\documentclass[letterpaper, 10 pt, conference]{ieeeconf} 
\IEEEoverridecommandlockouts
\usepackage{balance}
\usepackage{adjustbox}
\usepackage{color}
\usepackage[hidelinks,allcolors=blue,citecolor=blue,urlcolor=blue]{hyperref}

\usepackage{graphics} 
\usepackage{epsfig} 
\usepackage{mathptmx} 
\usepackage{times} 
\usepackage{amsmath} 
\usepackage{amssymb}  

\usepackage{url} 
\makeatletter
\let\NAT@parse\undefined
\makeatother
\usepackage{hyperref}  

\usepackage{wrapfig}

\usepackage{multirow}
\usepackage{tabularx,booktabs}

\usepackage{amsmath}

\newcommand{\fref}[1]{Fig.~\ref{#1}}
\newcommand{\sref}[1]{Section~\ref{#1}}
\newcommand{\tref}[1]{Table~\ref{#1}}

\usepackage{graphics} 
\usepackage{epsfig} 
\usepackage{amsmath} 
\usepackage{amssymb} 
\usepackage{color}
\usepackage{bm}
\usepackage{subcaption}
\usepackage{multirow}
\usepackage{array}
\usepackage{booktabs}
\usepackage{algorithm,algcompatible}
\usepackage{mathtools}
\usepackage{cite}
\usepackage{bbding}
\usepackage{cuted}
\definecolor{pointcolor}{RGB}{0,114,189}
\definecolor{linecolor}{RGB}{217,83,25}
\definecolor{planecolor}{RGB}{237,177,32}
\definecolor{spherecolor}{RGB}{119,172,48}
\definecolor{ellipsoidcolor}{RGB}{126,47,142}
\definecolor{cylindercolor}{RGB}{77,190,238}
\definecolor{conecolor}{RGB}{162,20,47}

\graphicspath{{figures/}}

\title{\huge \bf PhysORD: A Neuro-Symbolic Approach for Physics-infused Motion Prediction in Off-road Driving}

\author{Zhipeng Zhao, Bowen Li, Yi Du, Taimeng Fu, and Chen Wang
\thanks{Source code is available at \textcolor{blue}{\href{https://github.com/sair-lab/PhysORD}{https://github.com/sair-lab/PhysORD}}.}
\thanks{Zhipeng Zhao, Yi Du, Taimeng Fu, and Chen Wang are with Spatial AI \& Robotics (SAIR) Lab, Institute for Artificial Intelligence and Data Science, Department of Computer Science and Engineering, University at Buffalo, Buffalo, NY 14260, USA. {\tt\small Email: \{zhipengz, yid, taimengf, chenw\}@sairlab.org}}
\thanks{Bowen Li is with Robotics Institute, Carnegie Mellon University, Pittsburgh, PA 15213, USA. {\tt\small Email: bowenli2@andrew.cmu.edu}}
}

\begin{document}

\maketitle
\begin{abstract}

Motion prediction is critical for autonomous off-road driving, however, it presents significantly more challenges than on-road driving because of the complex interaction between the vehicle and the terrain. Traditional physics-based approaches encounter difficulties in accurately modeling dynamic systems and external disturbance. In contrast, data-driven neural networks require extensive datasets and struggle with explicitly capturing the fundamental physical laws, which can easily lead to poor generalization. By merging the advantages of both methods, neuro-symbolic approaches present a promising direction. These methods embed physical laws into neural models, potentially significantly improving generalization capabilities. However, no prior works were evaluated in real-world settings for off-road driving. To bridge this gap, we present PhysORD, a neural-symbolic approach integrating the conservation law, i.e., the Euler-Lagrange equation, into data-driven neural models for motion prediction in off-road driving. Our experiments showed that PhysORD can accurately predict vehicle motion and tolerate external disturbance by modeling uncertainties. 
The learned dynamics model achieves 46.7\% higher accuracy using only 3.1\% of the parameters compared to data-driven methods, demonstrating the data efficiency and superior generalization ability of our neural-symbolic method.


\end{abstract}
\section{Introduction}

Autonomous driving has transformed the way we envision transportation and mobility. However, the majority of these advancements \cite{grigorescu2020survey} have been concentrated on on-road driving scenarios, which operate within structured environments and predictable conditions \cite{lefkopoulos2020interaction,su2023double}. Off-road driving, on the other hand, represents a vastly different challenge \cite{triest2022tartandrive,sivaprakasam2024tartandrive} and is essential for many applications such as field exploration and rescue missions \cite{ hadsell2009learning, viswanath2021offseg}. It introduces complex interactions between the vehicle and diverse terrains, such as mud, gravel, and sand, which significantly affect the vehicle's motion and stability \cite{cai2023probabilistic, maheshwari2023piaug}. This complexity underscores the need for robust motion prediction that can navigate the unpredictability inherent in off-road environments.



Traditional physics-based models rely on motion formulas derived from fundamental principles, such as Newton's Laws of Motion \cite{newton1850newton}, for vehicle state prediction. While effective in their generalization ability, they struggle to accurately model the complex dynamic systems and external disturbances in off-road driving. 
Off-road conditions introduce various challenges, such as irregular terrain with bumps and load variations, which cause the system to exhibit highly nonlinear and nonstationary behavior.
Evolution models based on ideal assumptions, such as kinematic models \cite{ailon2005controllability, polack2018guaranteeing}, along with state estimation algorithms like the Kalman Filter \cite{kalman1960new} with simplified noise assumptions, face difficulties in predicting long-term motion due to external disturbances.

In parallel, leveraging the expressive neural networks, current data-driven methods \cite{hafner2019dream, tremblay2021multimodal, triest2022tartandrive} formulate the task as an end-to-end regression process and have shown promising progress in state prediction. However, these methods neglect physical laws and inherent constraints, such as conservation laws and symmetric structures. Thus, they demand extensive data for training and face challenges in generalizing to long-term prediction and unseen environments.

\begin{figure}[t]
    \centering
    \vspace{6pt}
    \includegraphics[width=\linewidth]{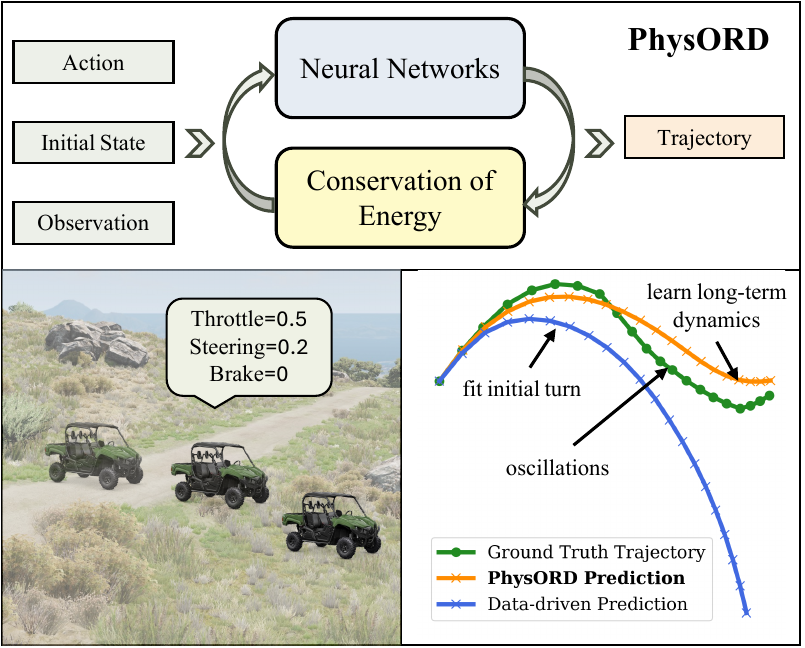}
    \caption{Overview of PhysORD and motion prediction task. Given the action, initial state, and observation, PhysORD predicts an accurate long-term trajectory (in the \emph{bottom right}) by combining neural methods and the conservation law.}
    \vspace{-4mm}
    \label{fig:intro}
\end{figure}

Recent research on neuro-symbolic approaches shows the potential to merge the strengths of both physics-based and data-driven methods. 
Researchers integrate symbolic reasoning from physical laws into neural networks, resulting in physics-infused neural models \cite{chen2018neural}. These models are designed to model processes such as the behavior of pendulums and mass-spring systems \cite{greydanus2019hamiltonian,saemundsson2020variational,duruisseaux2023lie}.
To preserve the inherent structure of dynamic systems throughout the learning process, these approaches establish a system of ordinary differential equations (ODEs), where unknown knowledge is parameterized using neural networks.
Previous works have delved into learning the Lagrangian \cite{lutter2018deep,saemundsson2020variational,havens2021forced,duruisseaux2023lie}, Hamiltonian \cite{greydanus2019hamiltonian}, and other non-linear differential equations \cite{raissi2019physics}. 
Despite these advancements, the performance of these methods relies on well-distributed data from simulations, with real-world implementation in robots facing significant external disturbance and uncertainty yet to be explored.






To close this gap, we extend the neuro-symbolic approaches to predict vehicle motion interacted with various terrain in real-world off-road scenarios. 
As shown in \fref{fig:intro}, we assume that a complex system consists of a known physical evolution process and an unknown external disturbance.
This allows us to utilize symbolic reasoning to model the known physical evolution process and use data-driven components to address the unknown uncertainty. 
Concretely, we model the off-road vehicle as a controlled Lagrangian system \cite{saemundsson2020variational} evolving on the Lie group \cite{duruisseaux2023lie} and governed by the principles of conservation and symmetry. 
Compared with the widely used kinematic models \cite{ailon2005controllability}, the Lagrangian mechanics not only capture the underlying physical laws but also provide a more general perspective into the dynamic system, especially for off-road scenarios where the kinematic process is extremely complex. 
Additionally, we estimate the force and potential energy with neural networks to encode uncertainty and learn the effects of external disturbance.



Our main contribution can be summarized as:
\begin{itemize}
\item We propose a \textbf{Phys}ics-infused motion prediction model for \textbf{O}ff-\textbf{R}oad \textbf{D}riving (PhysORD), which effectively combines the physical laws with neural networks.

\item Extensive experiments on the real-world TartanDrive dataset \cite{triest2022tartandrive} showed that PhysORD outperforms data-driven methods by \textbf{46.7\%} in accuracy using \textbf{3.1\%} of the parameters, exhibiting its data-efficient learning and generalization ability in long-term prediction.

\end{itemize}

\section{Related Works}
\label{sec:related_work}
Predicting vehicle motion relies on accurately modeling state changes over time. This section categorizes methods into physics-based, data-driven, and physics-infused neural networks, according to the way they describe the transitions.

\subsection{Physics-based Methods}
Physics-based methods model vehicle motion using fundamental physical laws, notably Newton's Laws of Motion \cite{newton1850newton}. These methods employ motion formulas derived from physical principles to predict future states.
A common example is the kinematic bicycle model \cite{ailon2005controllability, polack2018guaranteeing}, which simplifies a vehicle to a front-wheel-drive bicycle, determining the next state from the current state, acceleration, and steering angle. Despite their simplicity and efficiency, these models typically rely on precise sensor data and assume ideal conditions, limiting their effectiveness in real-world scenarios. To address these limitations, approaches such as the Kalman Filter \cite{kalman1960new} and Monte Carlo simulations \cite{mooney1997monte} have been developed to estimate states while accounting for noise and uncertainty. The Kalman Filter \cite{kalman1960new} uses a normal distribution to quantify uncertainty, combining predictions and measurements to refine state estimations. Monte Carlo simulations \cite{mooney1997monte} address uncertainty by sampling various input scenarios, and predicting various possible outcomes. 

Nevertheless, these methods often assume simplified noise models, like the Gaussian normal distribution, which may not adequately capture all external influences. Moreover, in complex systems or when internal parameters are unobservable externally, these models may fail, leading to decreased accuracy. Consequently, while offering a framework for incorporating uncertainty, physics-based models are generally more suited to short-term predictions and less reliable for complex or long-term scenarios.

\subsection{Data-driven Methods}
Data-driven approaches utilize end-to-end neural networks to model state evolution as a probabilistic process, learning from extensive datasets without relying on complex analytical equations. These methods leverage Recurrent Neural Networks (RNN) \cite{rumelhart1986learning}, including variations like LSTM \cite{hochreiter1997long} and GRU \cite{cho2014learning}, to address the challenge of retaining past input information over long prediction horizons. These networks incorporate hidden memory states to capture long-term dependencies in sequential data, making them well-suited for long-term motion prediction tasks. An example of their application is found in \cite{xu2019automated}, where LSTM demonstrated superior capability in capturing transitions from past states and actions to current states when tested on the Apollo autonomous driving platform across various vehicle types. Further building on RNN, model-based reinforcement learning methods have been developed and tested in both simulated \cite{tremblay2021multimodal} and real-world settings \cite{triest2022tartandrive}. These approaches focus on mapping multimodal observations into a latent space for accurate time-series forecasting, showcasing the versatility and effectiveness of data-driven strategies in predicting complex vehicle behaviors. However, their main drawback lies in the difficulty of capturing the underlying physical laws, which leads to limited generalization capabilities.

\subsection{Physics-infused Neural Networks}
Physics-infused Neural Networks combine both methods by integrating the physical law that governs the system into the neural network's learning process. Given the continuity of dynamic systems, a pioneer work, Neural Ordinary Differential Equation (ODE) \cite{chen2018neural} treats the evolution of the neural model's hidden state as a continuous process, rather than a discrete sequence of layers. By parameterizing the derivative of the hidden state, this method allows for the incorporation of ODE into neural networks. This integration motivates further research to learn Lagrangian \cite{lutter2018deep}, Hamiltonian \cite{greydanus2019hamiltonian}, and other general physics with non-linear differential equations \cite{raissi2019physics} by incorporating the structure introduced by the ODE into the learning. Particularly for systems governed by conservation laws, Hamiltonian Neural Networks (HNNs) \cite{greydanus2019hamiltonian} propose to parameterize the Hamiltonian function, enabling the network to learn energy dynamics directly from data, exemplified in the mass-spring and pendulum system. 

To further impose more physical property and structure constraints, such as symmetries, on the embedding space, Variational Integrator Networks (VINs) \cite{saemundsson2020variational} utilize a structure-preserving discretization method of variational integrators to derive discrete-time motion equations that maintain both geometric structure and physical laws.
Building on this, Forced Variational Integrator Networks (FVIN) \cite{havens2021forced} extend VINs for broader system applications, incorporating energy dissipation from external controls. The LieFVIN \cite{duruisseaux2023lie} further combines VINs and FVINs to learn a discrete time symplectic approximation for robot systems evolving on the Lie group and demonstrate the learned dynamics can be used in the control of drone in simulation. However, most of their applications are limited to simple physics or simulation environments. Extending these approaches to real-world ground vehicles remains challenging.

\section{Approach}

\label{sec:approach}
\begin{figure}[t]
    \centering
    \includegraphics[width=\linewidth]{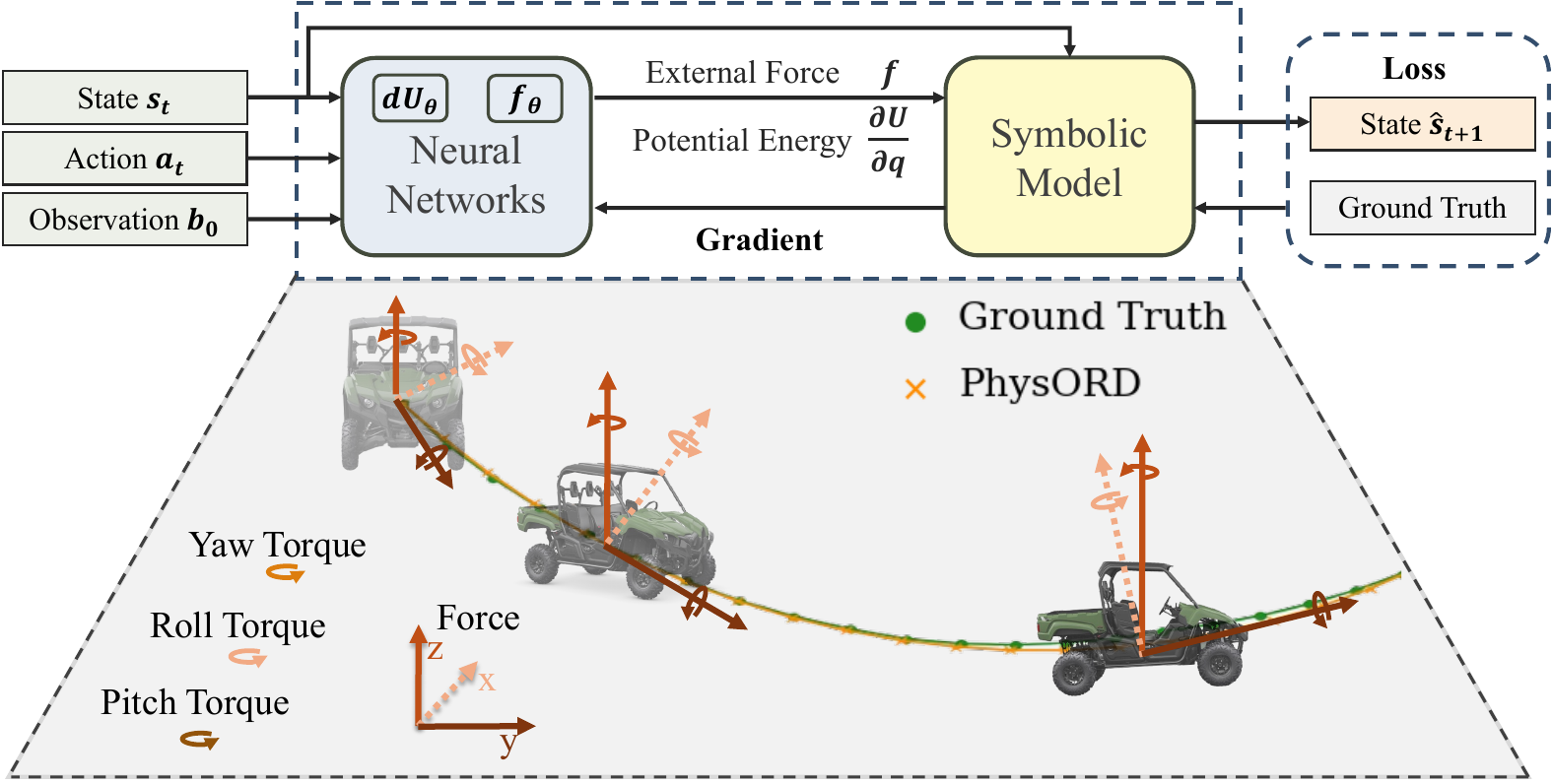}
    \caption{The Architecture of PhysORD. The neural networks contain two MLPs, $dU_\theta$ for potential energy prediction and $f_\theta$ for external force estimation. 
    Utilizing these estimated physical symbols, the symbolic model calculates the next state $\hat s_{t+1}$ from the current state $s_t$. The error between $\hat s_{t+1}$ and ground truth is backpropagated to optimize the MLPs.} \vspace{-5mm}
    \label{fig:network}
\end{figure}

\subsection{Overview}
The task of motion prediction for off-road vehicles can be formulated as follows: given an initial state $s_0$, an initial observation $b_0$, and a sequence of actions $\{a_0, a_1, \dots, a_{n-1}\}$, the objective is to forecast the sequence of future states $\hat{T} = \{\hat{s}_1, \hat{s}_2, \dots, \hat{s}_n\}$. 
To accomplish this, as illustrated in \fref{fig:network}, we introduce PhysORD, a neuro-symbolic model that integrates neural networks with a symbolic model to capture the dynamics of off-road vehicles effectively. For each time step $t$, PhysORD performs a forward prediction:
\begin{equation}
    \hat{s}_{t+1} = H(s_t, a_t, b_0),
\end{equation}
where $s_t=(q_t, \dot q_t)$ denotes the vehicle's state, comprising its pose $q_t$ and its linear and angular velocities $\dot q_t$. After $n$ iterations of forward predictions, we obtain states $\hat{T}$.

To accurately perform physics-infused forward prediction, PhysORD separates the known physical processes from less observable dynamics. The dynamic system conserves energy and is described with the discrete Euler-Lagrange equation. However, the potential energy and external force are challenging to quantify with precision. To bridge this gap, we utilize two neural networks ($dU_\theta$ and $f_\theta$) to predict these factors. The inferred values are subsequently integrated into the symbolic model, which computes the next state based on physical symbols. During the backpropagation, the symbolic model directs the gradient to refine the neural networks. We next first elaborate on the symbolic model grounded in the Euler-Lagrange equation, then introduce the neural networks to estimate potential energy and external force.


\subsection{Symbolic Model}
We employ \emph{Lagrangian and Hamiltonian mechanics} to develop a symplectic map that updates the state $s_t$, inspired by \cite{saemundsson2020variational, duruisseaux2023lie}. The evolution follows the conservation law and preserves the structure imposed by the constraints on Lie manifold \cite{wang2023pypose, zhan2023pypose}.
\emph{Hamilton mechanics} describes the dynamic system in terms of energy with the generalized coordinates $q$. According to \emph{Hamilton's Variational Principle}, the actual trajectory of a system is the one where the action integral is stationary for variations over a time interval $T$:
\begin{equation}
\label{eq:hamilton}
    \delta \int_{0}^{T} L(q(t), \dot{q}(t)) dt = 0,
\end{equation}
where $L(\cdot)$ is the Lagrangian function of the system. In scenarios involving external forces or controls $u(t)$, this principle extends to \emph{the Lagrange-d'Alembert Principle}:
\begin{equation} \label{eq:force_alembert}\fontsize{9.5pt}{11.4pt}\selectfont
    \delta \int_{0}^{T} L(q(t), \dot{q}(t)) \, dt + \int_{0}^{T} f_L(q(t), \dot{q}(t), u(t)) \cdot \delta q(t) \, dt = 0,
\end{equation}
where $f_L$ is the generalized force on the system. By discretizing its action integral with the variational integrators \cite{saemundsson2020variational}, \emph{the forced discrete Euler-Lagrange equation} can be obtained:
\begin{equation} \label{eq:discrete_euler_lagrange}
    D_2 L^d(q_{t-1}, q_t, h) + D_1 L^d(q_t, q_{t+1}, h) + f^{+} + f^{-} = 0.
\end{equation}
Here, $q_t=q(t)$ and $q_{t+1}=q(t+h)$ represent states at consecutive time steps with interval $h$, and $D_i$ is the partial derivative to the $i$-th argument. $f^{\pm}$ approximates the continuous Lagrangian force $f_L$ in a discrete setting.

For off-road vehicle dynamics, the generalized coordinates $q$, encompassing position $x$ and orientation $R$, evolve on the Lie group $\mathbb{SE}(3)$. The system's kinetic and potential energies define the Lagrangian $L$ as:
\begin{equation} \label{eq:lagrange}
L(x, R, v, \omega) = \frac{1}{2} v^T m v + \frac{1}{2} \omega^T J \omega - U(x, R),
\end{equation}
where $U$ is the potential energy, and $m$ and $J$ are the mass and inertia matrix, respectively.
By applying the discrete vehicle Lagrangian into \emph{the discrete Euler-Lagrange equation} ~\eqref{eq:discrete_euler_lagrange}, with a detailed derivation process shown in \cite{duruisseaux2023lie}, the state update equations from $t$ to $t+1$ can be obtained: 
\begin{subequations}\fontsize{8.95pt}{11.4pt}\selectfont
\begin{align}
&x_{t+1} = x_t + h v_t - (1 - \alpha) \frac{h^2}{m} \textcolor{blue}{\frac{\partial U_t}{\partial x_t}} + \frac{h}{m} R_t \textcolor{blue}{f_{t}^{x-}}, \label{eq:first} \\
&hS(J \omega_{t}) + hS(\textcolor{blue}{f_{t}^{R-}}) + (1 - \alpha)h^2S(\textcolor{blue}{\xi_t}) = Z_t J_d - J_d Z_t^T, \label{eq:second}\\
&R_{t+1} = R_t Z_t, \\
&mv_{t+1} = mv_t - (1 - \alpha) h \textcolor{blue}{\frac{\partial U_t}{\partial x_t}} - \alpha h \textcolor{blue}{\frac{\partial U_{t+1}}{\partial x_{t+1}}} + R_t \textcolor{blue}{f_{t}^{x-}} + R_{t+1} \textcolor{blue}{f_{t}^{x+}}, \label{eq:fourth} \\
&J\omega_{t+1} = Z_t^T J \omega_{t} + (1 - \alpha)h Z_t^T \textcolor{blue}{\xi_t} + \alpha h \textcolor{blue}{\xi_{t+1}} + Z_t^T \textcolor{blue}{f_{t}^{R-}} + \textcolor{blue}{f_{t}^{R+}}, \label{eq:fifth}
\end{align}
\end{subequations}
where $\alpha \in [0, 1]$, and $J_d$ is defined as $\frac{1}{2} \text{tr}(J) \mathbb{I}_3 - J$. The $S(\cdot)$ is the skew-symmetric matrix, and $S(\xi)=\frac{\partial U}{\partial R}^T R - R^T \frac{\partial U}{\partial R}$. 

Given the substantial mass and volume of the ground vehicle, precisely determining the potential energy derivative is challenging. Additionally, the complexity of the vehicle-terrain interaction complicates the calculation of forces. Therefore, 
we model those unknown (in blue) information with neural networks, denoted as $dU_\theta$ and $f_\theta$. These models will be detailed in Section \ref{method:nn} and are formulated as:
\begin{subequations}
\begin{align}
    \frac{\partial U_t}{\partial x_t}, \frac{\partial U_t}{\partial R_t}&=dU_\theta(x_t, R_t), \\
    f_{t}^{x\pm}, f_{t}^{R\pm}&=f_\theta(v_t, \omega_t, a_t, b_0).
\end{align}
\end{subequations}
The position $x_{t+1}$ is updated using Equation~\eqref{eq:first}. To determine the rotation matrix $Z_t$ which updates rotation $R$, Equation ~\eqref{eq:second} can be solved with a few steps of \emph{Newton’s method} detailed in \cite{duruisseaux2023lie}. Integrating these results, we obtain the derivatives of potential energy $\frac{\partial U_{t+1}}{\partial q_{t+1}}=dU_\theta(x_{t+1}, R_{t+1})$. This, in turn, influences the updates of linear and angular velocities, $v_t$ and $\omega_t$, in Equations~\eqref{eq:fourth} and \eqref{eq:fifth}, respectively.
Thus, we establish a comprehensive state update map:
\begin{equation}
    \{\hat{x}_{t+1},\hat{R}_{t+1}, \hat{v}_{t+1}, \hat{\omega}_{t+1}\} = F(x_t,R_t, v_t, \omega_t, dU_\theta, f_\theta).
\end{equation}


\subsection{Neural Networks}
\label{method:nn}
To address the gaps in physical information regarding the generalized forces $f^\pm$ and the potential energy $U$, we use two multi-layer perceptrons (MLPs) $dU_\theta$ and $f_\theta$ to learn these factors from the state, observation, and action inputs.

\subsubsection{Potential Energy MLP ($dU_\theta$)}
For a rigid body simplified to a point mass, calculating potential energy is straightforward. However, the large mass and volume of a car, combined with variations in the suspension and other components during movement, introduce significant errors. Moreover, the dynamic system's evolution utilizes the potential energy's partial derivatives, as indicated by the partial derivation $D_i$ in Equation ~\eqref{eq:discrete_euler_lagrange}. Instead of estimating the potential energy and calculating its differentials during optimization \cite{duruisseaux2023lie}, we implement a three-layer MLP to directly predict these derivatives. This approach generates a twelve-dimensional vector $dU$ from the vehicle's pose $q=(x,R)$:
\begin{equation}
    \frac{\partial U}{\partial q}=dU_\theta(q).
\end{equation}

The MLP configuration includes an input layer based on the pose, a hidden layer with ten units, and an output layer producing the derivatives $\frac{\partial U}{\partial x}$ and $\frac{\partial U}{\partial R}$, directly contributing to the dynamics' evolution. It is proved to produce a more accurate result as shown in Section \ref{res:abal}. 

\subsubsection{External Force MLP ($f_\theta$)}
The forces on off-road vehicles are significantly influenced by various factors, including driver inputs and terrain interactions, which are challenging to compute due to their complexity and uncertainty. Therefore, instead of using a predefined formula to compute force, where action inputs are scaled by learnable parameters \cite{duruisseaux2023lie}, we utilize an External Force MLP to infer forces from the combination of actions, states, and observations:
\begin{equation}
    f=f_\theta(\dot q, a, b),
\end{equation}
where $a$ includes throttle, steering, and brake, and $b$ measures the discrepancy between individual wheel speeds and the vehicle's overall speed, indicating terrain and environmental effects. These inputs, combined with vehicle velocities, are processed by the MLP. With layers configured as (13, 64), (64, 64), and (64, 6), the MLP predicts forces $f^{x}$ on the $x$, $y$, $z$ axes, and torques $f^{R}$ on pitch, roll, and yaw.


\subsection{Loss}
To optimize the neural networks, we employ a loss function that minimizes the discrepancy between predicted states $\hat{T} = \{\hat{s}_1, \hat{s}_2, \dots, \hat{s}_n\}$ with the ground truth. For the $x,v,\omega$, we calculate their Euclidean distance:
\begin{equation}
    L_{ED} = \frac{1}{n}\sum_{t=1}^{n} [(\hat{x_t} - x_t)^2 +(\hat{v}_t-v_t)^2 + \hat{\omega}_t - \omega_t)^2].
\end{equation}
For the rotation matrix, which operate in the $SO(3)$ space, we determine the relative rotation $R^{rel}_t$ between $\hat{R}_t$ and $R_t$ after normalization, then compute the geodesic distance:
\begin{equation}
\begin{aligned}
    L_{GD} &= \frac{1}{n}\sum_{t=1}^{n} [\cos^{-1}\left(\frac{\text{tr}(R^{rel}_t) - 1}{2}\right)]^2.
\end{aligned}
\end{equation}
The overall loss $L$ is defined as $L = L_{ED} + L_{GD}$.





\section{Experiments}
\label{sec:experiments}


\begin{table*}[t]\centering
\caption{Comparison of model prediction error. For each terrain type, the lowest errors in RMSE, Position distance $\bar{\rho}$, and Angular distance $\bar{\theta}$ across four methods are underlined. The lowest error across all terrain types is emphasized in bold.}\label{table:accuracy}
\resizebox{\textwidth}{!}{
\begin{tabular}{cc|ccc|ccc|ccc|ccc}\toprule
& &\multicolumn{3}{c|}{TartanDrive} &\multicolumn{3}{c|}{TartanDrive-variation} &\multicolumn{3}{c|}{Kalman Filter-NS} &\multicolumn{3}{c}{\textbf{PhysORD} (Ours)} \\
& &RMSE &Pos $\bar{\rho}$ &Ang $\bar{\theta}$ &RMSE &Pos $\bar{\rho}$ &Ang $\bar{\theta}$ &RMSE &Pos $\bar{\rho}$ &Ang $\bar{\theta}$ &RMSE &Pos $\bar{\rho}$ &Ang $\bar{\theta}$ \\\midrule
\multirow{7}{*}{\rotatebox[origin=c]{90}{Terrain Category}} &Gravel &1.4709 &0.7722 &0.0884 &1.4134 &0.7685 &0.0900 &1.3154 &0.9437 &0.3179 &\underline{0.7174} &\underline{0.5801} &\underline{0.0875} \\
&Plant &0.8578 &0.6916 &0.0964 &0.8402 &0.6738 &\underline{0.0887} &1.3345 &0.9674 &0.3319 &\underline{0.7152} &\underline{0.5763} &0.0911 \\
&Dirt &2.0216 &0.9346 &\underline{0.0877} &1.9176 &0.9022 &0.0930 &1.2707 &0.8997 &0.3064 &\underline{0.7681} &\underline{0.6065} &0.0895 \\
&Mud &2.4353 &1.1127 &0.0872 &2.3065 &1.0637 &\underline{0.0621} &1.3973 &0.9797 &0.3353 &\underline{0.8276} &\underline{0.6548} &0.0893 \\
&Puddle &0.6630 &0.5531 &\underline{0.0775} &0.7008 &0.5845 &0.1296 &1.1099 &0.7710 &0.2997 &\underline{0.6151} &\underline{0.5008} &0.0810 \\
&Rock &0.7060 &0.5896 &0.0748 &0.7148 &0.6183 &\underline{0.0732} &0.9974 &0.7390 &0.2427 &\underline{0.6290} &\underline{0.5194} &0.0755 \\
&Cement &\underline{0.7062} &0.6089 &0.0807 &0.7551 &0.6581 &0.0998 &1.3432 &1.0459 &0.2587 &0.7222 &\underline{0.5948} &\underline{0.0784} \\\midrule
\multicolumn{2}{c|}{\textbf{All Terrain}} & 1.3700 & 0.7706 & 0.0913 & 1.3094 & 0.7509 & 0.0936 & 1.2939 & 0.9329 & 0.3170 & \textbf{0.7297} & \textbf{0.5856} & \textbf{0.0893} \\
\bottomrule
\end{tabular}
}
\vspace{-10pt}
\end{table*}

This section describes the dataset used, evaluation methodologies, and the results obtained with our proposed approach.
\subsection{Dataset}
To evaluate the performance of our model in real-world off-road conditions, we conduct the experiments on the TartanDrive dataset \cite{triest2022tartandrive}.
This dataset encompasses approximately 2000 interactions of a Yamaha all-terrain vehicle (ATV) navigating through various complex terrains, such as driving through dense vegetation.
Such diversity presents both opportunities and challenges for learning off-road driving dynamics.
Additionally, TartanDrive serves as a benchmark for comparing our method against state-of-the-art data-driven approaches \cite{tremblay2021multimodal,hafner2019dream} in off-road motion prediction.

We divide the dataset into a training set and two evaluation sets, following the setting in \cite{triest2022tartandrive}. These evaluation sets are categorized by the level of difficulty, based on the average change in terrain height per second. The simpler set serves as the validation set during the training phase, while the more challenging set, featuring uneven terrain, is used for testing.

\subsection{Evaluation Details}
\label{exp:eval}
For a thorough evaluation of motion prediction accuracy, we employ the root mean squared error (RMSE) metric, as utilized in the TartanDrive experiments \cite{triest2022tartandrive}, and introduce two additional metrics.
RMSE quantifies the numerical discrepancies between the predicted pose vectors $\hat{q}$ and the actual pose vectors $q$. 
While it provides a measure of overall prediction accuracy, it combines errors in position and orientation, which may not fully represent their distinct physical implications. To address this, we introduce two specific metrics: Position distance and Angular distance, to offer a more intuitive physical evaluation of model performance.
\subsubsection{Position distance $\bar{\rho}$} This metric calculates the average Euclidean distance between predicted and actual positions, $\hat{x}$ and $x$, across all $N$ test sequences:
\begin{equation}
    \bar{\rho} = \frac{1}{N} \sum_{i=1}^{N} \|\hat{x_i} - x_i\|_2.
\end{equation}
\subsubsection{Angular distance $\bar{\theta}$} This metric calculates the mean geodesic distance between the predicted and actual orientations, evaluating the minimal rotation angle required to align the predicted rotation $\hat{R}$ with the actual rotation $R$.

As a supplement to RMSE, the two metrics introduce a complementary perspective that ensures a more balanced analysis.
Besides, to address the non-uniqueness of Euler angles and quaternions where different values may represent the same rotation, we standardize orientation representations of baseline models into rotation matrices for RMSE and angular distance calculations. This guarantees a fairer comparison of orientation prediction accuracy for different models. 

\subsection{Accuracy}
\label{exp:accuracy}
We evaluate the accuracy of both data-driven and neural-symbolic methods in the motion prediction task:

\subsubsection{TartanDrive} 
We select the best model presented in the TartanDrive dataset \cite{triest2022tartandrive} as the data-driven baseline and refer to it as the TartanDrive model.

\subsubsection{TartanDrive-variation} The TartanDrive model \cite{triest2022tartandrive} employs acceleration rather than the initial linear velocity, which our model uses. To enable a more thorough accuracy comparison, we revise the TartanDrive model to use the same input data as PhysORD, enabling comprehensive comparisons.

\subsubsection{Kalman Filter-NS} The Kalman Filter \cite{kalman1960new} is a classic state estimation algorithm with the ability to model uncertainty and Gaussian noise. Due to the absence of future measurements in our task, we cannot directly compare with it.
Therefore, we developed a Kalman Filter-based Neuro-Symbolic (Kalman Filter-NS) baseline model based on Gaussian noise assumption. 
In the prediction phase, we take the position and the rotation as the states and apply a constant velocity model as the state transition function.
The update phase incorporates external force effects through an MLP which predicts measurements from the state, action, and initial observation. These measurements, alongside the uncertainty evaluated during the prediction phase, refine the estimates of linear and angular velocity.

Each model predicts the future states with a 0.1-second time step, given initial states and a sequence of actions.
We assess model accuracy at the 20th step using RMSE, following \cite{triest2022tartandrive}, as well as Position distance $\bar{\rho}$, Angular distance $\bar{\theta}$ defined in \sref{exp:eval}. The evaluation was conducted across the entire evaluation dataset and also segmented by terrain type, with results presented in \tref{table:accuracy}.

Our PhysORD model demonstrates a notable 46.7\% improvement in RMSE over the data-driven neural method used in TartanDrive across all terrain \cite{triest2022tartandrive}, which is primarily due to more precise position predictions. This underscores PhysORD's effectiveness in integrating the prior physical knowledge of dynamic systems into motion prediction for off-road vehicles. The minor gains in angular distance highlight challenges in accurately capturing the pitch and roll dynamics due to significant uncertainty and noise for ground vehicles. By employing the linear velocity, rather than the acceleration used before, the TartanDrive-variation achieves a slight reduction in position error. However, its reliance on the data-driven approach limits its ability to understand the physical nature of motion, occasionally resulting in unrealistic trajectory predictions.

The Kalman Filter-NS method demonstrates a lower RMSE compared to neural-based approaches but does not improve the Position and Angular distance metrics. We observed that this is attributed to its ability to predict more stable trajectories by incorporating a kinematic model within the Kalman Filter, whereas the data-driven methods may produce unrealistic trajectories with a substantial error, significantly raising the RMSE. Nonetheless, this combination still has difficulty in modeling the complex off-road dynamics due to the simplification of physics and long-term uncertainty.

PhysORD outperforms other baseline models through a structure-preserving approach that infuses physical laws into neural networks. The integration of the Euler-Lagrange equation and MLPs is proven effective for off-road motion prediction with the improvement across all three metrics.

\subsection{Efficiency}
Efficiency is crucial for real-world robotics applications.
The ability of a model to dynamically adapt to new off-road environments through online learning depends greatly on training efficiency. Inference efficiency is essential for enabling real-time control and planning when the model is deployed on real robots. Thus, we evaluate both aspects.

\subsubsection{Training} 
\begin{figure}[t]
    \centering
    \includegraphics[width=\linewidth]{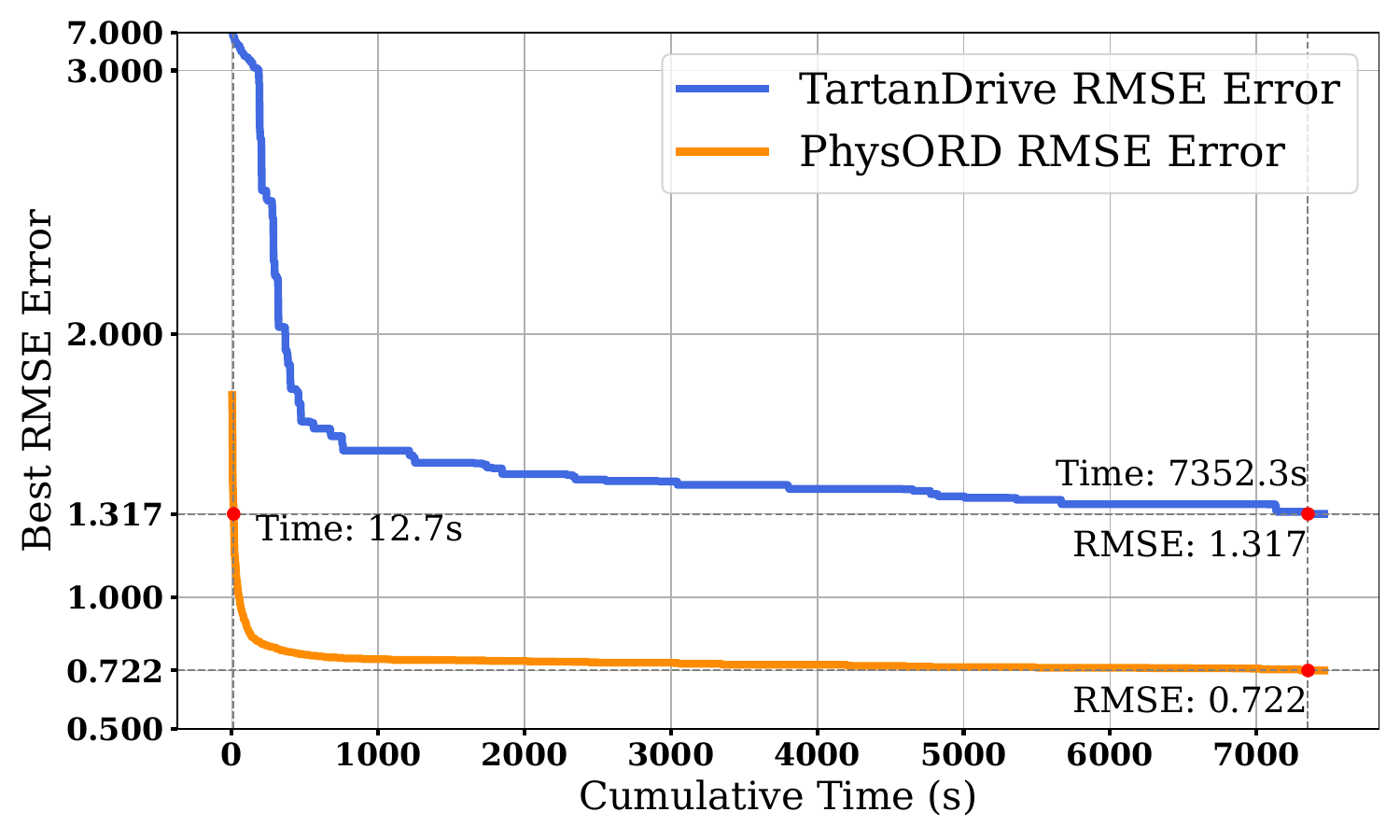}
    \caption{Comparison of training efficiency. This illustrates the lowest RMSE error achieved by both TartanDrive and PhysORD at various cumulative time points.} \vspace{-5mm}
    \label{fig:train_eff}
\end{figure}

We compared the best RMSE error achieved within the same training duration between the data-driven baseline and PhysORD, as depicted in \fref{fig:train_eff}. Infused with prior physical knowledge, PhysORD initiates with lower errors and undergoes rapid optimization, leading to a better final performance.
Notably, PhysORD attains TartanDrive's optimized performance in 0.17\% of the training time, as marked in \fref{fig:train_eff}. Besides, PhysORD's test error exhibits a smoother decline compared to the significant fluctuations observed with data-driven approaches. This demonstrates the symbolic model's capability to efficiently guide neural network optimization, overcoming the challenges faced by purely neural methods in learning complex patterns of off-road scenarios with a large search space.

\subsubsection{Inference} For inference efficiency, we measure the floating-point operations (FLOPs) required for a single forward pass through the model. \tref{table:infer_eff} shows that PhysORD consumes 23.4\% of the computation required by TartanDrive for a single inference. This significant reduction stems from PhysORD's much simpler neural network architecture, with 96.9\% fewer parameters than TartanDrive.

\begin{table}[h!]
\caption{Comparison of Inference Efficiency.}
\centering
\begin{tabular}{ c c c }
\toprule
  model & FLOPs & \# Params \\
\midrule
 TartanDrive \cite{triest2022tartandrive} & 543664 & 188.8k  \\ 
 PhysORD& \textbf{127100} & \textbf{5.8k} \\
\bottomrule
\end{tabular}
\vspace{-10pt}
\label{table:infer_eff}
\end{table}

In summary, our model outperforms neural network approaches in efficiency by effectively combining a symbolic model with neural networks that are simple and efficient.

\subsection{Long-term Prediction Generalization}
We evaluated the model's generalization capability by measuring its ability of learning short-term sequences to predicting long-term sequences. In Section \ref{exp:accuracy}, both the data-driven baseline and PhysORD were initially trained and tested using state data across 20 time steps. To evaluate their generalization ability, we retrain both models using sequences of 5 steps and evaluate their accuracy in predicting the state at the 20th time step, with results in \tref{table:general}.

\begin{table}[t]
\caption{Performance Comparison for Generalization.}
\centering
\begin{tabular}{ c c c c }
\toprule
  model & RMSE & Position distance & Angular distance\\
\midrule
 TartanDrive \cite{triest2022tartandrive} & 2.0321 & 1.5280 & 0.1329  \\ 
 PhysORD& \textbf{0.9514} & \textbf{0.7622} & \textbf{0.1115}\\
\bottomrule
\end{tabular}
\vspace{-12pt}
\label{table:general}
\end{table}

The performance gap between the models widens with fewer time steps of training data, increasing from a 24.0\% to a 50.1\% gap in Position distance. Compared to results in \tref{exp:accuracy} where models learn from 20 steps of data, the data-driven approach showed a significant increase in positional error by 98.3\%. In contrast, PhysORD's positional error increased by 30.2\%. Notably, PhysORD, trained on 5 steps of data, outperforms TartanDrive trained on 20 steps in both RMSE error and Position distance.

\subsection{Data-efficiency}
The amount of data for model training is another important aspect that affects real-world applications, especially for off-road driving where collecting data is challenging. We examine PhysORD's performance by testing various percentages of data for training: 1\%, 10\%, 50\%, 80\%, and 100\%. 

\begin{figure}[h!]
    \centering
    \includegraphics[width=\linewidth]{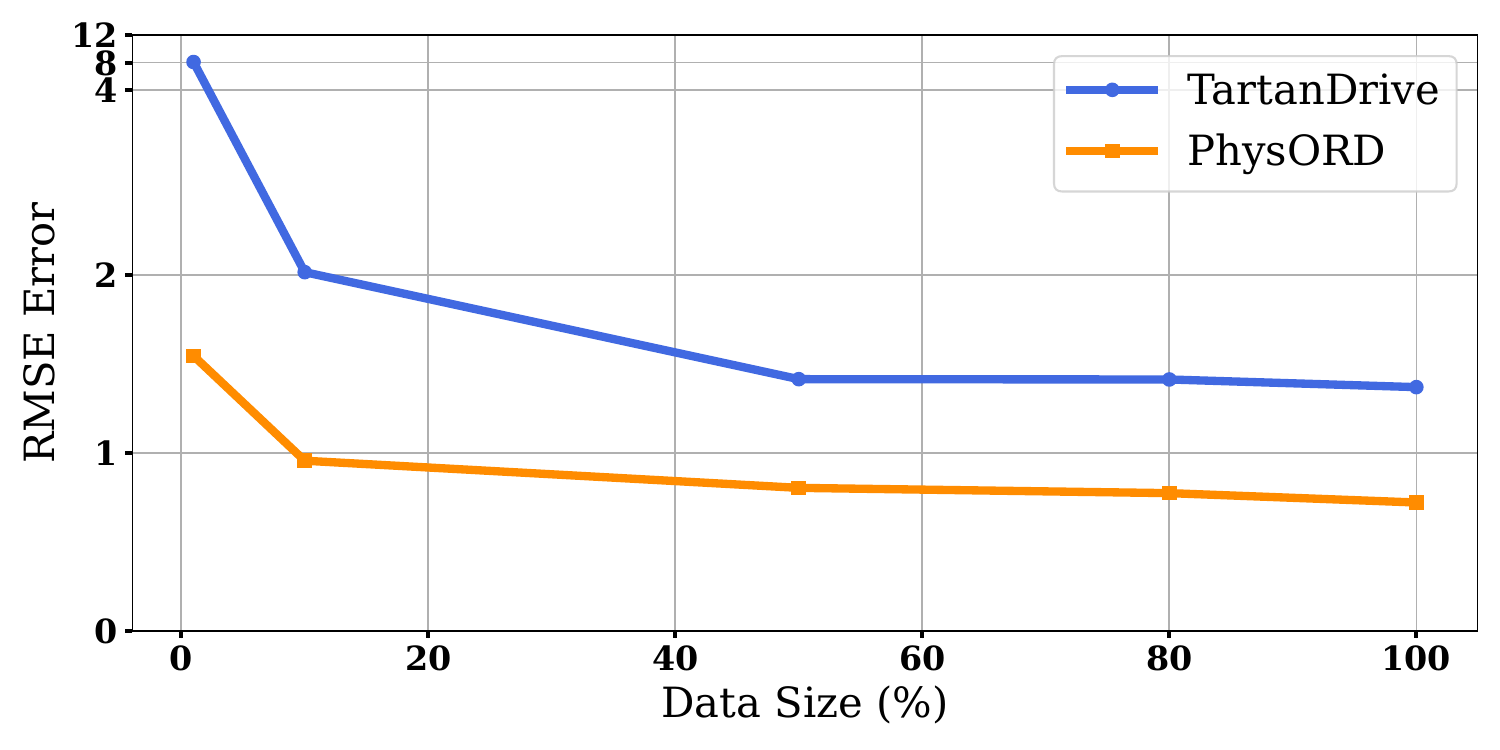}
    \caption{Accuracy comparison of data efficiency. The RMSE errors for TartanDrive and PhysORD when trained with various amounts of data from 1\% to 100\% of training set.} \vspace{-2mm}
    \label{fig:data_eff}
\end{figure}
\vspace{-0mm}

\begin{figure*}[t]
    \centering
    \includegraphics[width=\linewidth]{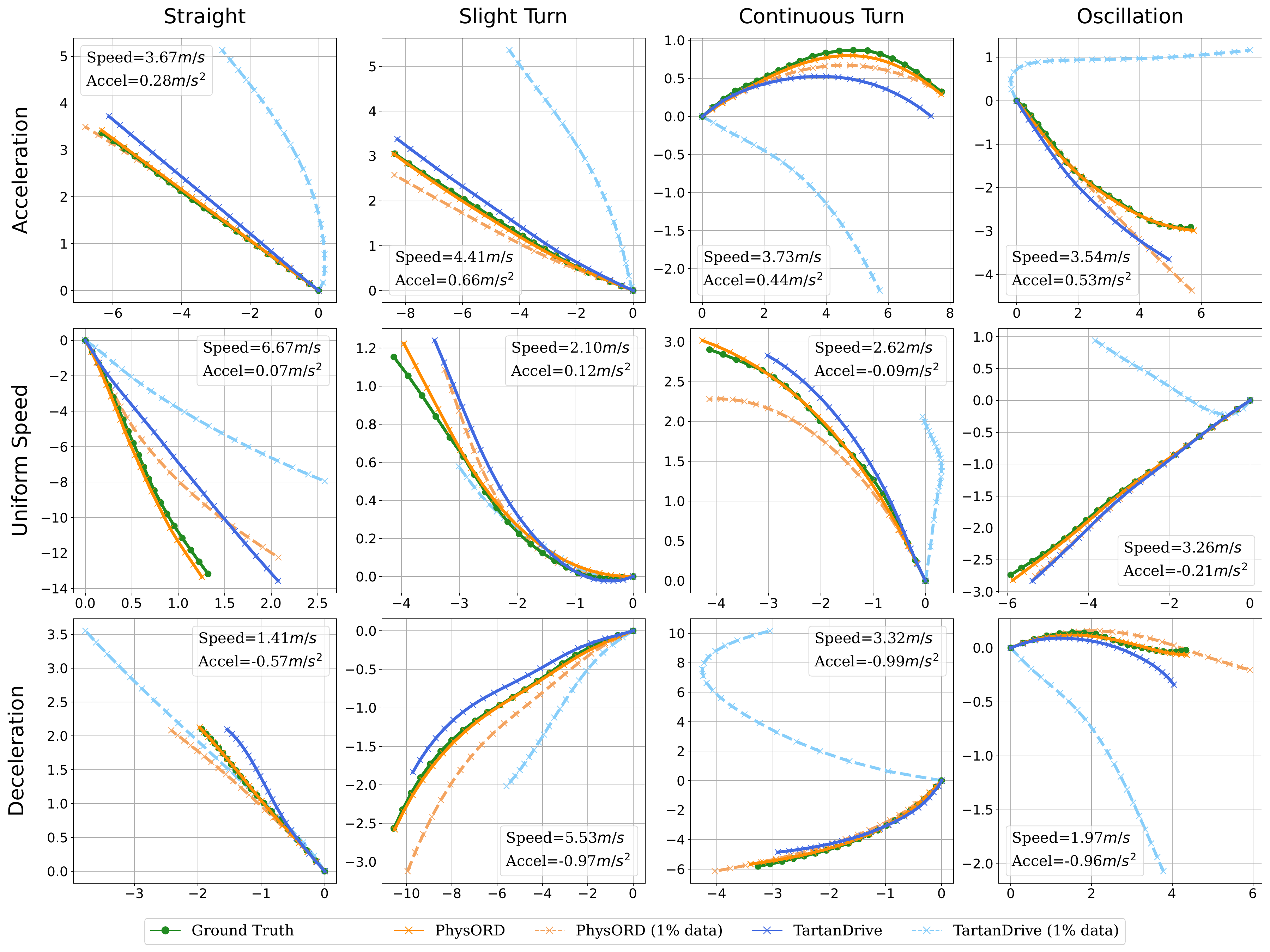}
    \caption{Qualitative analysis of PhysORD versus TartanDrive. Trajectories vary by speed change across rows and by motion type across columns, with speed and acceleration detailed in each subplot over these 20-time steps.} \vspace{-6mm}
    \label{fig:qualitative}
\end{figure*}

\fref{fig:data_eff} illustrates the RMSE errors for both the data-driven baseline and PhysORD at these data sizes. Even when data is reduced to 1\% of the training set, involving 5 continuous trajectories to form 647 sequences of 20-step data, the impact on performance is minimal, with RMSE errors increasing from 0.7297 to 1.4326. Conversely, TartanDrive's prediction error is magnified by nearly 5.9 times.

\subsection{Qualitative Results}
The evaluation set, capturing diverse challenges in off-road driving, consists of vehicle trajectories exhibiting various movements. These trajectories can be grouped into four categories based on motion type identified in the ground truth data: straight turn, slight turn, continuous turn, and oscillation. Within each category, speed variations including acceleration, uniform speed, and deceleration are observed. We present visual comparisons of 20-step predicted trajectories covering these distinct movements in \fref{fig:qualitative}, highlighting the performance improvement by our PhysORD.

PhysORD excels in long-term motion prediction, offering more accurate trajectories. While the data-driven model captures basic patterns like straight movement and turns from extensive data, it struggles with precise off-road dynamics, especially as the number of steps increases. This issue becomes more severe under complex conditions such as non-uniform velocity and oscillation as shown in the last column of \fref{fig:qualitative}. However, PhysORD generates stable and accurate trajectories that closely match the ground truth by utilizing the Euler-Lagrangian motion equation and effectively learning incomplete knowledge from neural networks. For challenging motions like oscillation with deceleration, PhysORD outperforms the data-driven approach, showcasing its superior adaptability in capturing the trajectories.

\subsection{Ablation Study}
\label{res:abal}

\begin{table}[h!]
\centering
\caption{Ablation Study.}
\setlength\tabcolsep{4.5pt}
\begin{tabular}{ c | c c c| c c c  }
\toprule
 model & Physic & F-NN & dU-NN & RMSE & $\hat{\rho}$ & $\hat{\theta}$\\
\midrule
 Ours-Phys & \XSolidBrush & \Checkmark & \Checkmark & 18.3962 & 2.9616 & 1.9539\\
 Ours-F & \Checkmark & \XSolidBrush & \Checkmark & 1.3855 & 1.0581 & 0.2685 \\
 Ours-U & \Checkmark & \Checkmark & \XSolidBrush & 0.7472 & 0.5891 & 0.0902 \\ 
 PhysORD & \Checkmark & \Checkmark & \Checkmark &\textbf{0.7297} & \textbf{0.5856} & \textbf{0.0893}\\
\bottomrule
\end{tabular}
\vspace{-15pt}
\label{table:abla}
\end{table}

We next conduct an ablation study to evaluate the impact and contribution of each component within our proposed PhysORD framework. PhysORD integrates a symbolic model with two neural networks: an external force MLP (F-NN) and a potential energy MLP (dU-NN). We developed three variants of our PhysORD for this analysis:

\subsubsection{Ours-Phys} Without the symbolic model grounded in Euler-Lagrangian Equations, Ours-phys learns the state updates in a pure neural approach using F-NN and dU-NN.

\subsubsection{Ours-F} While the F-NN in our proposed method predicts the external force directly from the state, observation, and action input, Ours-F adopts LieFVIN's approach \cite{duruisseaux2023lie} of defining the force equation by action and learning the physical equation's parameter.

\subsubsection{Ours-U}
Unlike dU-NN in PhysORD, which directly estimates partial potential energy, Ours-U, following the approach in \cite{saemundsson2020variational, duruisseaux2023lie}, predicts total potential energy and calculates its differentials during optimization.



As illustrated in \tref{table:abla}, the symbolic model is crucial within our neuro-symbolic approach. Without underlying physical knowledge, the purely neural approach struggles to predict how external factors affect off-road vehicle dynamics accurately. The comparison between Ours-F and PhysORD highlights that our F-NN can more effectively capture the impact of external actions, leading to improved performance in off-road driving. The impact of varying potential energy MLPs appears marginal, primarily attributed to the relatively minor Z-axis movement of ground vehicles.
\section{Conclusions and Future Work}
\label{sec:conclusion}
This paper introduces PhysORD, a neuro-symbolic model that infuses the Euler-Lagrange equation with neural networks for motion prediction in off-road driving scenarios. By modeling the vehicle as a controlled Lagrangian system and using MLPs to estimate forces and potential energy, PhysORD addresses the challenges posed by complex dynamics and environmental uncertainties. 
Experiments on the TartanDrive dataset show that PhysORD outperforms data-driven methods, improving prediction accuracy by 46.7\% while using 96.9\% fewer parameters. This reduction in parameters notably enhances both training and inference efficiency. Moreover, PhysORD exhibits the ability to learn from limited data and generalize from short-term learning to long-term prediction.
Future work will focus on integrating environmental data, such as forward-facing images and terrain height maps, to further improve performance in real-world off-road vehicle control and planning.

\section*{Acknowledgements}

This work was in part supported by the ONR award N00014-24-1-2003 and DARPA grant DARPA-PS-23-13. Any opinions, findings, conclusions, or recommendations expressed in this paper are those of the authors and do not necessarily reflect the views of the ONR or DARPA.
The authors also wish to express their gratitude for the generous gift funding provided by Cisco Systems Inc.

\balance
\bibliographystyle{IEEEtran}
\bibliography{references}

\end{document}